\documentclass[lettersize,journal]{IEEEtran}
\usepackage{amsmath,amsfonts}
\usepackage{algorithmic}
\usepackage{algorithm}
\usepackage{array}
\usepackage{tabularx}
\usepackage{booktabs}
\usepackage{amssymb}
\usepackage{multirow}
\usepackage[caption=false,font=normalsize,labelfont=sf,textfont=sf]{subfig}
\usepackage{textcomp}
\usepackage{stfloats}
\usepackage{url}
\usepackage{verbatim}
\usepackage{graphicx}
\usepackage{tikz}
\usepackage{cite}
\usepackage{chapterbib}
\usepackage{xcolor}
\usepackage[hidelinks,hypertexnames=false]{hyperref}
\IfFileExists{orcidlink.sty}{\usepackage{orcidlink}}{\newcommand{\orcidlink}[1]{\href{https://orcid.org/#1}{\textsuperscript{\textcolor[HTML]{A6CE39}{\scriptsize iD}}}}}
\newcolumntype{Y}{>{\raggedright\arraybackslash}X}
\newcolumntype{L}[1]{>{\raggedright\arraybackslash}p{#1}}
\newcolumntype{R}[1]{>{\raggedleft\arraybackslash}p{#1}}
\newcommand{\pmetric}{Precision@(F1=1, IoU$\geq$0.5)}
\newcommand{\nmetric}{N-acc}
\newcommand{\pmetricsym}{$\mathcal{P}$}
\newcommand{\nmetricsym}{$\mathcal{N}$}
\newcommand{\arxivsubmissionnotice}{This work has been submitted to the IEEE Transactions on Multimedia for possible publication.\\Copyright may be transferred without notice, after which this version may no longer be accessible.}
\newif\ifarxivnoticeprinted

\begin{document}

\AddToHook{shipout/foreground}{%
  \ifarxivnoticeprinted\else
    \begin{tikzpicture}[remember picture,overlay]
      \node[anchor=north west,align=left,font=\scriptsize,text width=.95\paperwidth,inner sep=0pt] at ([xshift=4pt,yshift=-2pt]current page.north west) {\arxivsubmissionnotice};
    \end{tikzpicture}%
    \global\arxivnoticeprintedtrue
  \fi
}

\title{UniRef-UAV: A Multimodal Benchmark for Universal Referring in UAV Imagery}

\author{Haibin~Tian\orcidlink{0009-0009-0753-3428}, Huichao~Xie\orcidlink{0009-0009-7117-3834}, Xuelin~Qian\orcidlink{0000-0001-8049-7288},~\IEEEmembership{Member,~IEEE}, Ruitao~Lu\orcidlink{0000-0002-7527-4298}, Junwei~Han\orcidlink{0000-0001-5545-7217},~\IEEEmembership{Fellow,~IEEE}, and Dingwen~Zhang\orcidlink{0000-0001-8369-8886},~\IEEEmembership{Member,~IEEE}%
\thanks{Haibin Tian and Huichao Xie contributed equally to this work.}%
\thanks{Haibin Tian, Huichao Xie, Xuelin Qian, and Dingwen Zhang are with the School of Automation, Northwestern Polytechnical University, Xi'an 710072, China (e-mail: haibintian@foxmail.com; xiehuichao@mail.nwpu.edu.cn; xlqian@nwpu.edu.cn; zdw2006yyy@nwpu.edu.cn). Corresponding author: Dingwen Zhang.}%
\thanks{Ruitao Lu is with the College of Missile Engineering, Rocket Force University of Engineering, Xi'an 710038, China (e-mail: lrt19880220@163.com).}%
\thanks{Junwei Han is with the School of Artificial Intelligence, Chongqing University of Posts and Telecommunications, Chongqing 400065, China (e-mail: jhan@nwpu.edu.cn).}%
}

\markboth{IEEE Transactions on Multimedia}%
{UniRef-UAV: A Multimodal Benchmark for Universal Referring in UAV Imagery}

\maketitle

\begin{abstract}
Unmanned aerial vehicles (UAVs) increasingly rely on visual grounding capabilities to localize task-relevant targets from diverse instructions in complex aerial scenes. Existing referring expression comprehension (REC) benchmarks and methods, however, are largely built around text-only queries and single-object outputs, which limits their applicability to practical UAV scenarios involving reference images, multimodal instructions, absent targets, and multiple valid target instances. To address this gap, we introduce \emph{Universal Referring}, a generalized UAV referring task that jointly expands the query modality and the output cardinality. We construct \emph{UniRef-UAV}, a multimodal benchmark that supports text-only, image-only, and text+image queries with modality-dependent target cardinality, where text-only and text+image queries admit no-target, single-target, and multi-target grounding while image-only queries focus on existence-aware single-instance grounding. It also provides in-domain and cross-domain evaluation protocols for visual-query generalization. We further present \emph{UAV-URNet}, a detection-style baseline that maps heterogeneous queries into a shared query space and predicts variable-size target sets through set prediction. Extensive experiments show that UAV-URNet provides a stable and reproducible baseline with more consistent no-target discrimination and a more lightweight, reproducible implementation than large general-purpose MLLMs. Additional domain analysis, query-representation analysis, and ablation studies demonstrate that multimodal queries help reduce visual-query ambiguity and promote a more unified query--target alignment space. The annotations, visual query crops/images, train/validation/test splits, evaluation scripts, and baseline code will be made publicly available to facilitate reproducible research.
\end{abstract}

\begin{IEEEkeywords}
UAV vision, multimodal referring, universal referring, visual grounding, variable-cardinality detection.
\end{IEEEkeywords}

\section{Introduction}

\IEEEPARstart{U}{nmanned} aerial vehicles (UAVs) have become an important mobile sensing platform for security patrols~\cite{dudukcu2023uavsurvey}, emergency search and rescue~\cite{lasalandra2024uavemergency,MoDeTrack}, traffic monitoring~\cite{ChenH14}, and industrial inspection~\cite{YangZNYLYCWFH26}. 
Related visual systems have also expanded toward large-scale scene reconstruction and open-vocabulary scene querying~\cite{CoSurfGS,KDDistributedDynamic3DGS,LangSurfTPAMI}, but practical UAV operations still require interfaces that ground task instructions to concrete visual targets.
Beyond autonomous flight and navigation, these applications increasingly require UAVs to interpret task-level instructions and associate them with specific visual targets in complex aerial scenes. 
For example, a rescue UAV must localize the indicated victim or area before approaching or delivering supplies, while a surveillance UAV must identify the particular person, vehicle, or object of interest before tracking or triggering an alarm. 
Such requirements highlight referring understanding as a key capability for UAV-based multimodal scene understanding, where natural-language descriptions, reference images, or other multimodal cues must be grounded to the correct scene entities to support reliable downstream decision-making and action.

\begin{figure*}[!t]
\centering
\includegraphics[width=\textwidth]{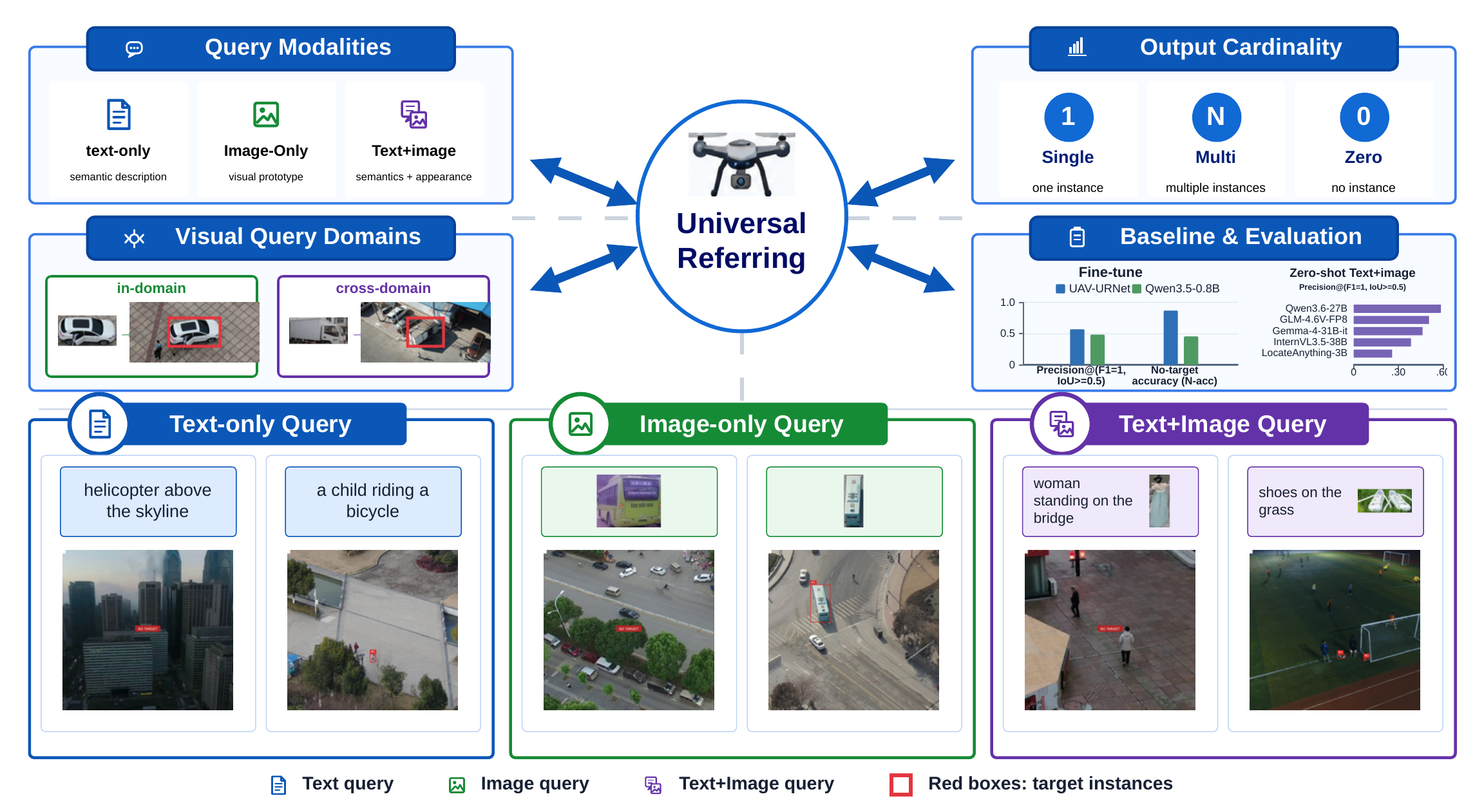}
\caption{Overview of Universal Referring and the UniRef-UAV benchmark.
UniRef-UAV jointly expands UAV referring along query modality and modality-dependent target cardinality: text-only and text+image queries support no-target, single-target, and multi-target grounding, while image-only queries support existence-aware single-instance grounding. Representative UAV cases, in-domain/cross-domain visual-query settings, and baseline evaluation results are shown, where red boxes denote target instances.}
\label{fig:universal-referring-query-cases}
\end{figure*}

Referring in UAV imagery differs substantially from conventional referring expression comprehension~\cite{grounding_survey}. 
Aerial scenes usually cover large spatial extents and contain many distractor instances with similar semantics or appearances, making target disambiguation highly challenging. 
Moreover, referred targets in UAV imagery often occupy only a few pixels and exhibit low resolution, weak texture, and ambiguous boundaries, further complicating fine-grained recognition and localization~\cite{SkyFind}.
These visual difficulties are coupled with limitations in the query modality. Language is effective for conveying high-level semantics, intent, and spatial relations, but it may be insufficient or cumbersome for rare, visually complex, or hard-to-verbalize targets. 
Visual references, in contrast, provide concrete appearance cues, yet lack the abstraction and compositional expressiveness of text.
Moreover, UAV tasks naturally exhibit variable target cardinality, where a single query may correspond to no object, one object, or multiple object instances. 
Collectively, query-modality limitations and variable target cardinality challenge the standard REC setting, which typically relies on text-only queries and assumes a single referred target~\cite{refcoco,refcocog,SkyFind}, and motivate a more flexible multimodal referring framework for UAV scenarios.

To address this issue, we formulate \textbf{Universal Referring}, a more general UAV referring paradigm with \textbf{multimodal query inputs} and \textbf{flexible-cardinality outputs}, as illustrated in Fig.~\ref{fig:universal-referring-query-cases}. 
Universal Referring supports text-only, image-only, and text+image queries, and adopts modality-dependent output cardinality: text-only and text+image queries allow zero-to-many targets, while image-only queries are restricted to existence-aware single-instance grounding due to prototype ambiguity.

To instantiate this task, we introduce \textbf{UniRef-UAV}, a multimodal benchmark built from 22 public UAV datasets spanning urban traffic, low-altitude flight, and open-field environments.
On top of these images, we manually curate three query forms, namely text-only, image-only, and text+image queries, to cover both semantic and appearance-driven referring needs.
UniRef-UAV further adopts flexible target-cardinality annotations, including no-target, single-target, and multi-target cases, resulting in more than 150K query samples.
As a benchmark, it provides both in-domain and cross-domain evaluation settings, enabling systematic analysis of how well models localize referred targets when the visual query comes from the same or different UAV data domains.

Beyond the UniRef-UAV benchmark, we further develop the \textbf{UAV Universal Referring Network (UAV-URNet)} as a baseline framework tailored to the proposed task.
Unlike conventional REC models that regress one bounding box for each query, Universal Referring requires the model to determine whether the referred target exists and to predict a target set with variable cardinality.
UAV-URNet addresses this requirement with a detection-style set prediction framework.
It extracts semantic representations for both scene images and visual queries, projects heterogeneous query modalities into a shared query space, and performs query-conditioned candidate matching to produce a variable number of bounding boxes. 
By decoupling candidate box generation from cross-modal matching, UAV-URNet can flexibly integrate text-only, image-only, and text+image queries while explicitly handling no-target, single-target, and multi-target cases without substantial architectural modification. 
Experiments on UniRef-UAV show that this framework provides a stable, reproducible, and extensible baseline for multimodal universal referring in UAV scenarios.

The main contributions of this paper are summarized as follows:
\begin{enumerate}
  \item We identify the distinctive challenges of referring understanding in UAV imagery and formulate \textbf{Universal Referring}, a generalized task that supports multimodal query inputs and modality-dependent output cardinality.

  \item We build \textbf{UniRef-UAV}, a multimodal universal referring benchmark for aerial scenes. It supports three query forms, i.e., text-only, image-only, and text+image queries, covers diverse UAV scenarios and output cardinalities, and provides both in-domain and cross-domain evaluation protocols.

  \item We propose \textbf{UAV-URNet}, a detection-style baseline framework for the proposed task. The model accommodates different query modalities and predicts variable-size target sets, providing a simple, reproducible, and extensible baseline for UAV multimodal referring research.
\end{enumerate}

\section{Related Work}

\begin{table*}[!t]
\centering
\footnotesize
\setlength{\tabcolsep}{3pt}
\renewcommand{\arraystretch}{1.12}
\caption{Comparison of existing GREC-related datasets and UniRef-UAV.}
\label{tab:dataset-comparison}
\begin{tabular*}{\textwidth}{@{\extracolsep{\fill}}lcccccccccc@{}}
\toprule
Dataset & Domain & Img & Obj & Expr & Avg Len. & Expr Type & Avg Res. & Avg O/I & MQ & GREC \\
\midrule
gRefCOCO\cite{GREC} & Life & 20.0K & 60.7K & 258.8K & 4.98 & Text & 593$\times$485 & 10.12\% & $\times$ & $\checkmark$ \\
VQDv1\cite{VQDv1} & Life & 122.8K & 604.9K & 621.5K & 6.31 & Text & 578$\times$484 & 18.48\% & $\times$ & $\checkmark$ \\
D3\cite{D3} & Life & 10.6K & 18.5K & 422 & 6.31 & Text & 746$\times$613 & 9.92\% & $\times$ & $\checkmark$ \\
SkyFind\cite{SkyFind} & UAV & 35.6K & 352.9K & 1.0M & 27.19 & Text & 2192$\times$1305 & 0.74\% & $\times$ & $\times$ \\
UniRef-UAV & UAV & 48.5K & 210.3K & 157.0K & 4.73 & Text/Image & 1455$\times$860 & 1.30\% & $\checkmark$ & $\checkmark$ \\
\bottomrule
\end{tabular*}
\par\vskip 0.5em
\begin{minipage}{\textwidth}
\footnotesize\raggedright
\emph{Note:} Img, Obj, and Expr denote the numbers of images, annotated objects, and referring expressions, respectively. Avg Len. denotes the average expression length, Avg Res. denotes the average image resolution, and Avg O/I denotes the average object-to-image area ratio. MQ indicates multimodal query support, and GREC indicates whether variable-cardinality referring annotations are provided.
\end{minipage}
\end{table*}

\subsection{Classical Referring Expression Comprehension in UAV Perspective}

SkyFind~\cite{SkyFind} represents an important step toward UAV-based REC by introducing a large-scale aerial benchmark with textual referring expressions. 
It highlights key challenges in aerial referring, including small-object localization, background suppression, and long-range spatial relation modeling. Nevertheless, SkyFind still follows the conventional REC paradigm: the query is limited to text, and each expression is mainly associated with a single existing target. These limitations motivate a more general UAV referring formulation that supports multimodal queries and flexible output cardinality.

\subsection{Generalized Referring Expression Comprehension}

Representative GREC datasets, including gRefCOCO~\cite{GREC}, VQDv1~\cite{VQDv1}, and D3~\cite{D3}, relax the single-target assumption by allowing a textual query or description to correspond to zero, one, or multiple object instances. 
As summarized in Table~\ref{tab:dataset-comparison}, however, these datasets remain centered on natural images and text-only queries, and therefore do not address the UAV-specific multimodal referring setting studied in this paper.

\subsection{Multimodal Queried Grounding}

Recent multimodal query-based detection and grounding methods have begun to combine textual prompts with visual exemplars, aiming to compensate for the limited descriptive capacity of text alone in fine-grained categories, long-tail objects, and visually ambiguous concepts. MQ-Det~\cite{MQDet} extends language-queried detection by incorporating visual exemplars into category representations. 
T-Rex2~\cite{TRex2} further unifies text prompts, visual prompts, and their combination within an open-set detector. By aligning textual and visual prompts through contrastive learning, it enables flexible switching among different prompt forms during detection.
Language-vision prompt pre-training has also been explored for low-data instance segmentation~\cite{UPLVP}, further reflecting the utility of coupling language and visual cues under limited supervision.

Nevertheless, in MQ-Det and T-Rex2, multimodal queries mainly serve as class-level prompts or enhanced category prototypes: visual exemplars are used to define or refine a category concept, and the detector is expected to retrieve all instances belonging to that category. In contrast, Universal Referring aims to localize a query-specific target set determined by the joint semantics of text, image, or their combination.
Image-only referring is also related to one-shot object detection, where a visual support example guides target localization~\cite{FSODFeatureReweighting,RepMet,OS2D,CoAE,AIT}. However, these methods are typically category-oriented and evaluated by AP or mAP, whereas Universal Referring treats the query image as a visual semantic prototype and further requires existence-aware grounding under a set-level cardinality protocol.

\section{Universal Referring Task Definition}

\subsection{Problem Formulation}

We reformulate visual referring as a unified, modality-agnostic semantic grounding problem. Given an image \(\mathbf{I} \in \mathbb{R}^{H \times W \times 3}\) and a query \(\mathbf{Q}\), the model is required to localize all image regions that satisfy the query condition. The output is a set of bounding boxes \(\mathbf{B} = \{\mathbf{b}_i\}_{i=1}^{|\mathbf{B}|}\), where each \(\mathbf{b}_i\) denotes one target instance grounded by \(\mathbf{Q}\).

Unlike conventional referring expression comprehension, our formulation explicitly incorporates both the \textbf{query modality} and the \textbf{output cardinality} into the task definition.

\textbf{Input Query.}
We consider three query types:
\[
\mathbf{Q} \in \{ \mathrm{T},\; \mathrm{V},\; \mathrm{T+V} \}
\]

where $\mathrm{T}$ is a text description (text-only), $\mathrm{V}$ is a reference image (image-only), and $\mathrm{T{+}V}$ is a reference image with text (text+image).

\textbf{Output Set.}
The model outputs a set of boxes whose valid cardinality depends on the query modality:

\[
|\mathbf{B}| \in \mathcal{C}(\mathbf{Q}), \quad
\mathcal{C}(\mathbf{Q}) =
\begin{cases}
\{0,1,\dots,N\}, & \mathbf{Q}\in\{\mathrm{T},\mathrm{T{+}V}\},\\
\{0,1\}, & \mathbf{Q}=\mathrm{V}.
\end{cases}
\]

Classical REC and GREC can be viewed as special cases of Universal Referring. 
When $\mathbf{Q}=\mathrm{T}$ and $|\mathbf{B}|=1$, the task reduces to conventional REC; when $\mathbf{Q}=\mathrm{T}$ and $|\mathbf{B}|\in\mathcal{C}(\mathrm{T})$, it corresponds to GREC-style text-conditioned variable-cardinality grounding. Universal Referring further extends this space by introducing image-only and text+image queries, thereby jointly generalizing the query modality and modality-dependent output cardinality of visual referring.

We assume that grounding semantics can be represented independently of query modality, while different modalities provide different levels of semantic specificity. 
As shown in Table~\ref{tab:semantic-specification}, the three query modalities differ in their semantic specification, boundary definition, hierarchy control, cardinality, and grounding type. Text-only queries explicitly specify semantic concepts and set boundaries, enabling flexible grounding with variable cardinality. In contrast, image-only queries provide visual semantic prototypes without explicit semantic boundary specification, making multi-instance grounding fundamentally ambiguous. Therefore, image-only grounding is formulated as a prototype-based existence and localization problem, where the model retrieves the instance that best matches the query image under supervision-induced grounding semantics. 
Text+image queries combine symbolic semantic specification with visual appearance grounding, providing the strongest semantic disambiguation capability.

\begin{table}[htbp]
\centering
\footnotesize
\setlength{\tabcolsep}{3pt}
\renewcommand{\arraystretch}{1.12}
\caption{Comparison of the three query modalities.}
\label{tab:semantic-specification}
\begin{tabularx}{\columnwidth}{lYYY}
\toprule
Property & Text-only & Image-only & Text+Image \\
\midrule
Semantic specification & Explicit & Implicit prototype & Disambiguated \\
Semantic boundary & Well-defined & Ambiguous & Well-defined \\
Hierarchy control & Explicit & Implicit & Explicit \\
Cardinality & 0/1/Many & 0/1 & 0/1/Many \\
Grounding type & Set & Prototype & Hybrid \\
\bottomrule
\end{tabularx}
\end{table}

Universal Referring treats query modality as part of the task semantics rather than a superficial input variation. It exposes a fundamental coupling between semantic expressiveness and output cardinality, requiring different query forms to be unified within one grounding formulation while respecting their distinct grounding assumptions.

\subsection{Evaluation Metrics}
Following the GREC evaluation protocol~\cite{GREC,grounding_survey}, we evaluate Universal Referring by set-level exact correctness. For target-present samples, predictions are matched to ground-truth boxes through one-to-one assignment under IoU$\geq0.5$, and a sample is counted as correct only when the matched prediction set achieves $\mathrm{F1}=1$, where
\[
\mathrm{F1} = \frac{2\mathrm{TP}}{2\mathrm{TP} + \mathrm{FP} + \mathrm{FN}}.
\]
We report \textbf{\pmetric}, defined as
\[
\mathrm{Precision@}(\mathrm{F1}=1,\mathrm{IoU}\geq0.5)
= \frac{1}{|\mathcal{D}|}\sum_{j\in\mathcal{D}}\mathbb{I}[\mathrm{F1}_j=1],
\]
where $\mathcal{D}$ denotes the evaluated sample set. For no-target samples, exact correctness is defined by empty-set prediction rather than by the box-matching F1 formula, and we report \textbf{\nmetric} on the no-target subset $\mathcal{D}_0$:
\[
\mathrm{N\text{-}acc}
= \frac{1}{|\mathcal{D}_0|}
\sum_{j\in\mathcal{D}_0}
\mathbb{I}[|\hat{\mathbf{B}}_j|=0].
\]
This evaluation protocol jointly measures localization accuracy and output-cardinality correctness. For image-only queries, where multi-target cases are excluded by construction, the evaluation reduces to existence-aware single-instance grounding. In the following tables, we use \pmetricsym{} and \nmetricsym{} to denote \pmetric{} and \nmetric, respectively.

\section{Dataset Construction}

\subsection{Construction Necessity}

To assess the dataset gap behind UAV universal referring, we conduct a cross-dataset training and evaluation study. Specifically, we train two representative variable-output referring models, SimVG~\cite{ming2024simvg} and MMGroundingDINO~\cite{GroundingDINO,zhao2024open}, on three different data sources: gRefCOCO~\cite{GREC} for generalized referring in ground-view natural images, SkyFind~\cite{SkyFind} for conventional text-based UAV REC, and the proposed UniRef-UAV. All models are evaluated on the same UniRef-UAV text+image test set. This comparison reveals whether existing ground-view GREC data or text-only UAV REC data can be directly transferred to the proposed UAV universal referring setting.

\begin{table}[htbp]
\centering
\footnotesize
\setlength{\tabcolsep}{4pt}
\renewcommand{\arraystretch}{1.08}
	\caption{Cross-dataset evaluation on the UniRef-UAV test set.}
\label{tab:construction-necessity-results}
\begin{tabular*}{\columnwidth}{@{\extracolsep{\fill}}llcc@{}}
\toprule
Model & Train & \pmetricsym(\%) & \nmetricsym(\%) \\
\midrule
S & SkyFind & 1.71 & 7.17 \\
S & gRefCOCO & 10.59 & 5.13 \\
S & U-TI$^{-}$ & 18.47 & 3.12 \\
\midrule
D & SkyFind & 2.18 & 12.14 \\
D & gRefCOCO & 11.61 & 29.29 \\
D & U-TI$^{-}$ & 46.64 & 50.97 \\
D$^{*}$ & U-TI$^{+}$ & \textbf{54.64} & \textbf{93.53} \\
\bottomrule
\end{tabular*}
\par\vskip 0.5em
\begin{minipage}{\columnwidth}
\footnotesize\raggedright
\emph{Note:} S and D denote SimVG ViT-L and MM-GroundDINO-Tiny, respectively. U-TI$^{-}$ and U-TI$^{+}$ denote UniRef-UAV text+image training split without and with image query inputs, respectively. $^{*}$ denotes the modified version extended according to the baseline design in Section~\ref{sec:baseline} to support image query inputs.
\end{minipage}
\end{table}

As shown in Table~\ref{tab:construction-necessity-results}, models trained on existing datasets transfer poorly to the UniRef-UAV test set. Training on gRefCOCO improves over SkyFind in most cases, indicating that variable-cardinality supervision is helpful. However, both ground-view GREC data and conventional text-based UAV REC data remain far behind models trained directly on UniRef-UAV. In particular, incorporating image-query supervision further improves the MM-GroundDINO baseline, demonstrating the benefit of visual query information in the proposed setting.

These results suggest that the performance gap is not merely caused by the domain shift between ground-view and aerial imagery. More importantly, it reflects a mismatch between existing referring datasets and the requirements of UAV universal referring, where multimodal queries and variable target cardinalities must be handled jointly. Therefore, existing public referring datasets are insufficient for this task, and a dedicated multimodal UAV grounding benchmark such as UniRef-UAV is necessary.

\subsection{Data Sources}

UniRef-UAV is built from images collected from 22 publicly available UAV-related datasets. These sources cover a wide range of aerial scenarios, including urban traffic monitoring, low-altitude flight, and open-field environments, providing diverse visual content for the proposed universal referring setting. After frame sampling, uniform subsampling, visual deduplication, and manual quality filtering, UniRef-UAV contains 48,470 unique images. The complete source list and image filtering protocol are provided in the \emph{Data Source Details} section of the supplementary material.

\subsection{Annotation Process}
\label{sec:query-modal-design}

UniRef-UAV is annotated through a unified human-in-the-loop process designed around three requirements: multimodal query construction, variable-cardinality supervision, and image-level split control. Annotators first create text-only queries by selecting an object instance or an object set in the target image, drawing tight bounding boxes, and writing a natural-language expression that matches the selected targets in both semantics and cardinality. Text+image queries are then built by attaching a visual query example to the text annotation. Specifically, an in-domain visual query is cropped from a nearby frame in the same source environment and corresponds to the same object instance, whereas a cross-domain visual query is collected from an external web image with the same semantic meaning. Image-only queries are derived from the text+image annotations by removing the textual condition, while ambiguous multi-target samples are excluded.

To reflect practical UAV operations, UniRef-UAV further includes no-target queries, where the queried object is plausible in the scene category but absent from the current target image. Annotation quality is controlled through batch-wise mutual verification between annotator groups, and batches with an error rate above the preset threshold are returned for correction. Finally, the dataset is split at the target image level under a 7:1:2 train/validation/test ratio, with source datasets, query modalities, and output-cardinality buckets balanced as much as possible. The complete annotation and splitting protocol is provided in the \emph{Annotation and Split Details} section of the supplementary material.

\subsection{Dataset Statistics}

After the above annotation process, UniRef-UAV contains \textbf{48,470 images} and \textbf{156,987 task-ready base queries}. Among these queries, 110,869 correspond to a single target, 34,616 correspond to multiple targets, and 11,502 are no-target queries. The detailed target-cardinality distribution is reported in Table~\ref{tab:cardinality-distribution}.

\begin{table}[htbp]
\centering
\caption{Target-cardinality distribution of base queries.}
\label{tab:cardinality-distribution}
\footnotesize
\setlength{\tabcolsep}{4pt}
\begin{tabular*}{\columnwidth}{@{\extracolsep{\fill}}lcr@{}}
\toprule
Type & \#Queries & Ratio \\
\midrule
no target & 11,502 & 7.33\% \\
single target & 110,869 & 70.62\% \\
multi target & 34,616 & 22.05\% \\
\textbf{Total} & \textbf{156,987} & \textbf{100\%} \\
\bottomrule
\end{tabular*}
\end{table}

\begin{table}[htbp]
\centering
\caption{Statistics of the training, validation, and test splits.}
\label{tab:dataset-split-stats}
\footnotesize
\setlength{\tabcolsep}{3pt}
\renewcommand{\arraystretch}{1.08}%
\begin{tabular*}{\columnwidth}{@{\extracolsep{\fill}}lrrrr@{}}
\toprule
\multicolumn{5}{@{}l}{\emph{Base queries}} \\
\midrule
Setting & Train & Val. & Test & Total \\
\midrule
text-only & 64,103 & 15,582 & 25,849 & 105,534 \\
text+image & 34,253 & 6,878 & 10,322 & 51,453 \\
Total & 98,356 & 22,460 & 36,171 & 156,987 \\
\addlinespace[0.6ex]
\midrule
\multicolumn{5}{@{}l}{\emph{Unique images}} \\
\midrule
Setting & Train & Val. & Test & Total \\
\midrule
text-only & 33,927 & 4,847 & 9,696 & 48,470 \\
text+image & 32,992 & 4,570 & 9,358 & 46,920 \\
Total & 33,927 & 4,847 & 9,696 & 48,470 \\
\bottomrule
\end{tabular*}
\end{table}

\begin{figure}[htbp]
\centering
\subfloat[]{%
\includegraphics[width=0.48\linewidth]{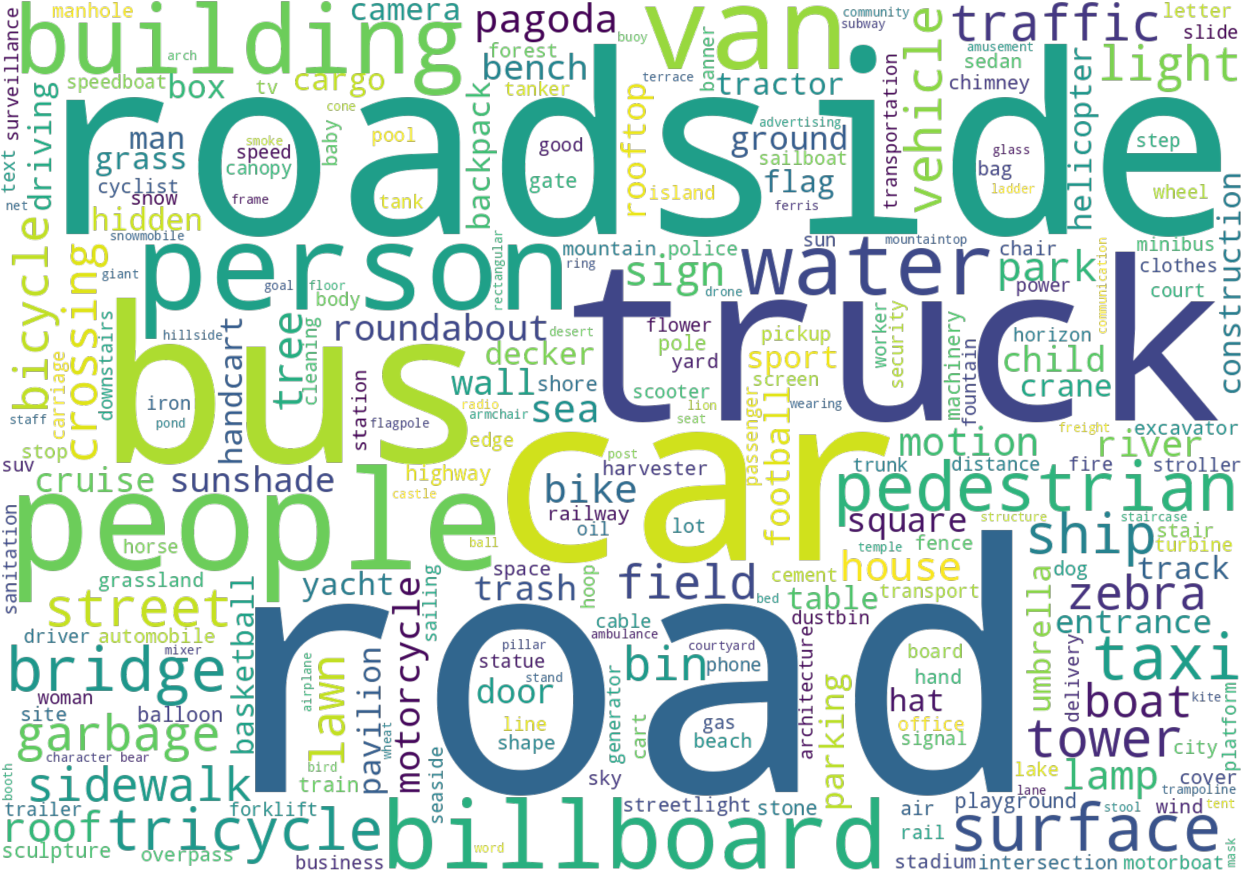}
\label{fig:noun-wordcloud-v2}}
\hfill
\subfloat[]{%
\includegraphics[width=0.48\linewidth]{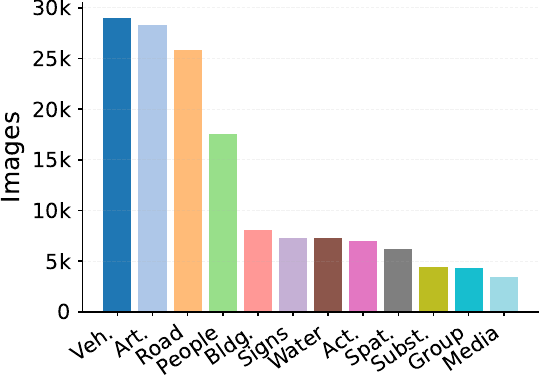}
\label{fig:image-number-by-category-v2}}
\par\medskip
\subfloat[]{%
\includegraphics[width=0.48\linewidth]{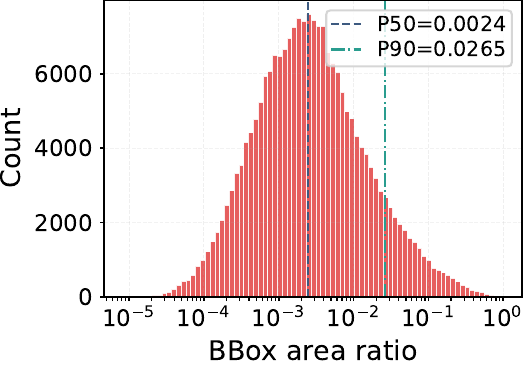}
\label{fig:box-size-ratio}}
\hfill
\subfloat[]{%
\includegraphics[width=0.48\linewidth]{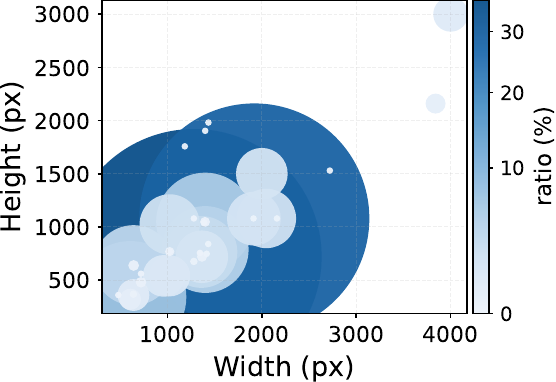}
\label{fig:image-resolution-bubble}}
\caption{Dataset statistics of UniRef-UAV. (a) Noun-frequency word cloud. (b) Image-category distribution. (c) Bounding-box area ratio distribution. (d) Image-resolution distribution.}
\label{fig:noun-frequency-stats}
\end{figure}

\textbf{Image resolution distribution.} We visualize the image resolutions in Fig.~\ref{fig:image-resolution-bubble}. UniRef-UAV covers a broad range of resolutions, from low-resolution frames to high-resolution aerial images, while maintaining a relatively balanced distribution. Common video resolutions, such as 1080p and 720p, account for a large portion of the dataset.

\textbf{Text-query noun statistics.} We analyze the semantic coverage of text queries through noun-frequency statistics. 
As shown in Fig.~\ref{fig:image-number-by-category-v2} and Fig.~\ref{fig:noun-wordcloud-v2}, high-frequency nouns mainly correspond to typical UAV targets, such as vehicles, pedestrians, and buildings, while the overall distribution spans diverse semantic categories.

\textbf{Instance statistics.}
We further analyze the distribution of annotated bounding boxes in terms of both their quantity and spatial scale. Specifically, we compute the ratio between each bounding box area and its corresponding image area (Fig.~\ref{fig:box-size-ratio}), and categorize all object instances into \emph{small}, \emph{medium}, and \emph{large} according to the standard COCO area definitions~\cite{COCO}. The resulting size statistics are summarized in Table~\ref{tab:box-size-category}.
As shown in Fig.~\ref{fig:box-size-ratio}, the bounding-box area ratios exhibit a near bell-shaped distribution, with most instances concentrated in a small relative-area range, i.e., approximately 0.001--0.010 of the image area. This observation is consistent with the small-object nature of UAV imagery. Meanwhile, Table~\ref{tab:box-size-category} shows that COCO-defined small objects do not dominate the dataset. This is mainly because many images in UniRef-UAV have high resolutions: even objects with small relative areas may exceed the COCO pixel-area threshold for small objects and are therefore categorized as medium or large.

\textbf{Visual-query domain distribution.} 
For the text+image query setting, we further report the distribution of in-domain and cross-domain visual queries under different target cardinalities. As shown in Table~\ref{tab:text-plus-image-cardinality-domain}, the ratio between in-domain and cross-domain queries is approximately 13:7 across no-target, single-target, and multi-target cases. This design supports a detailed evaluation of model behavior under both domain-consistent and domain-shifted visual queries.

\begin{table}[htbp]
\centering
\caption{Bounding-box scale distribution under different query settings.}
\label{tab:box-size-category}
\small
\resizebox{\columnwidth}{!}{%
\begin{tabular}{lrrrr}
\toprule
Setting & \#Boxes & Small & Medium & Large \\
\midrule
text--only & 136,257 & 40,028 (29.38\%) & 60,904 (44.70\%) & 35,325 (25.93\%) \\
text+image & 74,076 & 20,655 (27.88\%) & 32,637 (44.06\%) & 20,784 (28.06\%) \\
\bottomrule
\end{tabular}
}
\end{table}

\begin{table}[htbp]
\centering
\caption{Visual-query domain distribution of text+image queries under different target cardinalities.}
\label{tab:text-plus-image-cardinality-domain}
\small
\resizebox{\columnwidth}{!}{%
\begin{tabular}{lrrr}
\toprule
Target Cardinality & In-domain & Cross-domain & Total \\
\midrule
0 / no target & 1,482 (62.32\%) & 896 (37.68\%) & 2,378 \\
1 / single target & 24,348 (65.45\%) & 12,852 (34.55\%) & 37,200 \\
many / multi target & 8,000 (67.37\%) & 3,875 (32.63\%) & 11,875 \\
\midrule
\textbf{Total} & \textbf{33,830 (65.75\%)} & \textbf{17,623 (34.25\%)} & \textbf{51,453} \\
\bottomrule
\end{tabular}
}
\vskip 0.5em
\begin{minipage}{\columnwidth}
\footnotesize \textit{Note:} Percentages are computed within each target-cardinality row.
\end{minipage}
\end{table}

\section{Universal Referring Baseline}

To validate the feasibility of Universal Referring in UAV imagery, UAV-URNet is designed as a unified baseline framework that handles multiple query modalities and variable-sized outputs (Fig.~\ref{fig:baseline-model-abc}). The goal of UAV-URNet is not to pursue state-of-the-art performance, but to provide a \textbf{simple, reproducible, and extensible reference implementation}. Concretely, UAV-URNet is designed to: (i) support different query modalities within a single architecture and (ii) naturally accommodate predictions with varying cardinality.

\subsection{Detection-Style Grounding Architecture}

\begin{figure*}[!t]
\centering
\includegraphics[width=\textwidth]{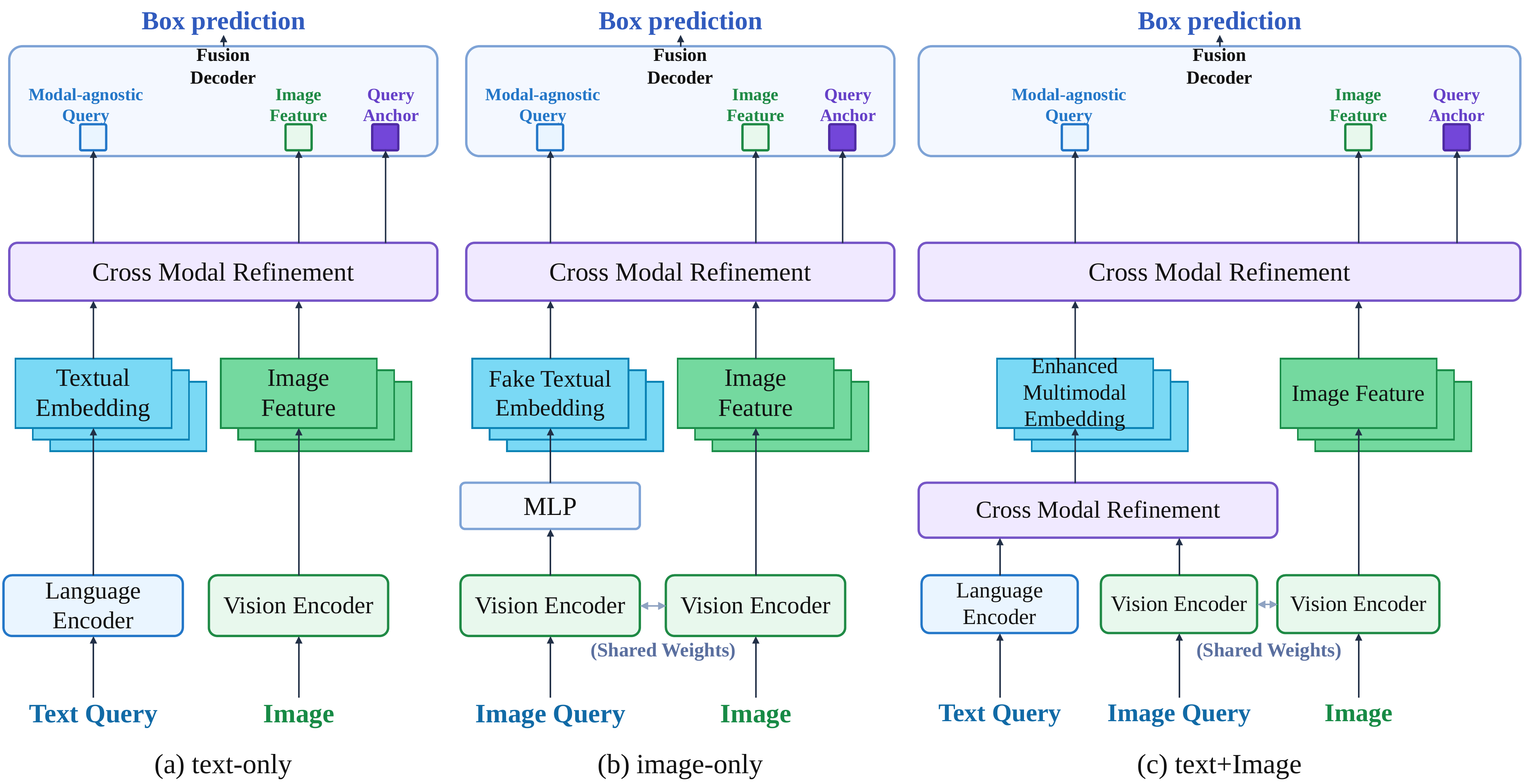}
\caption{Overview of UAV-URNet under three query types. Text-only queries follow the original GroundingDINO text-conditioned pipeline. Image-only queries are encoded by the shared vision encoder and projected through a lightweight MLP into the text-token dimension as pseudo query tokens. Text+image queries use the visual-query features to refine the text tokens, and the resulting multimodal query tokens are fed into the original cross-modal refinement and detection decoder.}
\label{fig:baseline-model-abc}
\setcounter{subfigure}{0}%
\refstepcounter{subfigure}\label{fig:baseline-model-a}%
\refstepcounter{subfigure}\label{fig:baseline-model-b}%
\refstepcounter{subfigure}\label{fig:baseline-model-c}%
\end{figure*}

Our baseline implementation builds upon GroundingDINO~\cite{GroundingDINO}, a classical detection-style grounding model that follows the \textit{2+2 architecture}~\cite{grounding_survey}. This architecture consists of four main components: an image encoder, a text query encoder, a cross-modal interaction module, and a fusion decoder. The image and query encoders first extract modality-specific features, which are then fused through cross-modal interactions and decoded into bounding box predictions (see Fig.~\ref{fig:baseline-model-abc}(a)).

We adopt this detection-style set prediction paradigm because it naturally matches the requirements of Universal Referring. First, set prediction allows the model to output zero, one, or multiple boxes, which is essential for variable-cardinality grounding. Second, decoupling candidate generation from query-conditioned matching makes it easier to incorporate heterogeneous query representations without redesigning the box prediction pipeline. Third, the framework operates on query embeddings rather than modality-specific query formats, making it suitable for extending text-conditioned grounding to image-only and text+image queries.

\subsection{Unified Query Representation}
\label{sec:baseline}

To support different query modalities within a shared backbone, we adopt a unified query representation strategy rather than introducing new task-specific heads. Concretely, we map text-only, image-only, and text+image queries into a common query embedding space that can be directly consumed by the existing detection encoder and decoder.

\textbf{Text-only Queries}.
For text-only queries (Fig.~\ref{fig:baseline-model-a}), UAV-URNet directly follows the original GroundingDINO text-conditioned grounding pipeline, using the text embedding as the query representation without introducing additional modules.

\textbf{Text+Image Queries}.
For multimodal queries that contain both text and image inputs (Fig.~\ref{fig:baseline-model-c}), we extend the text query formulation into a multimodal query formulation by fusing the textual query with the visual query representation. Specifically, we encode the query image using the same vision backbone as the target image, and refine the textual representation via the same cross-attention layers that take the query image features as keys and values, analogous to the original cross-attention mechanism, where information from the target image and the text query is fused. This operation fuses the two query streams and produces a multimodal query representation. The refined multimodal query is then fed into the cross-modal refinement module, where it participates in the same cross-modal interactions as in the text-only case.
This design preserves the text-driven inference process of the original framework, while allowing the query image to provide additional disambiguating context with minimal changes to the original architecture.

\textbf{Image-only Queries}.
For image-only queries (Fig.~\ref{fig:baseline-model-b}), the query image is encoded using the same vision encoder as the target image. The resulting feature representation is then projected through a lightweight MLP to match the dimensionality of the textual embedding space. The projected feature is treated as a \emph{pseudo-token} that serves as the query embedding during cross-modal matching.
This simple adaptation allows image-based queries to be seamlessly integrated into the existing grounding architecture without modifying the detection backbone or the decoding process.

\subsection{Training Strategy}

During training, we freeze both the visual encoder and the text encoder, and optimize only the cross-modal interaction layers, the projection MLPs, and the detection decoder. This strategy preserves the representation power of large-scale pretrained encoders while substantially reducing training cost and implementation complexity. 
It is important to emphasize that the purpose of this baseline is \textbf{task validation rather than architectural optimization}. We avoid introducing specialized modules or auxiliary supervision so that the framework remains transparent and easy to reproduce. As such, the proposed baseline provides a stable and extensible starting point for future research on universal referring in UAV imagery.
We keep the original GroundingDINO set-prediction losses unchanged, including the matching and box regression objectives, and only adapt the query-representation pathway for visual and multimodal inputs.

\section{Experiments}

\subsection{Benchmark Results}

We first evaluate representative methods from four families: 2+1 single-target transformers~\cite{TransVG,clip-vg,ming2024simvg}, detection-style grounding models~\cite{GroundingDINO,Mdetr}, general-purpose MLLMs~\cite{qwen36_27b,GLM-4.6V,MiMo-VL,InternVL35,Gemma4}, and grounding-oriented MLLMs~\cite{LLaVA-Grounding,GLaMM,Ferret-v2,Kosmos-2,Griffon,Rex-Omni,Rex-Seek,LocateAnything}. The MLLM prompting and post-processing protocol is provided in the \emph{MLLM Prompting and Post-processing Protocol} section of the supplementary material. The detailed results under text-only, image-only, and text+image query settings are summarized in Table~\ref{tab:main-existing-methods}.

\begin{table*}[!t]
\centering
\footnotesize
\caption{Performance of representative methods under text-only, image-only, and text+image query settings.}
\label{tab:main-existing-methods}
\setlength{\tabcolsep}{3pt}
\renewcommand{\arraystretch}{1.08}
\begin{tabular*}{\textwidth}{@{\extracolsep{\fill}}lcccccc@{}}
\toprule
Model & \multicolumn{2}{c}{Text-only} & \multicolumn{2}{c}{Image-only} & \multicolumn{2}{c}{Text+Image} \\
\cmidrule(lr){2-3}\cmidrule(lr){4-5}\cmidrule(lr){6-7}
 & \pmetricsym & \nmetricsym & \pmetricsym & \nmetricsym & \pmetricsym & \nmetricsym \\
\midrule
\multicolumn{7}{c}{\textbf{2+1  single-target transformer}} \\
\cmidrule(lr){1-7}
TransVG~\cite{TransVG} & 0.034 & 0.000 & -- & -- & -- & -- \\
CLIP-VG~\cite{clip-vg} & 0.052 & 0.000 & -- & -- & -- & -- \\
SimVG~\cite{ming2024simvg} & 0.045 & 0.002 & -- & -- & -- & -- \\
\midrule
\multicolumn{7}{c}{\textbf{Detection-style (2+2) transformer}} \\
\cmidrule(lr){1-7}
MMGroundingDINO~\cite{zhao2024open} & 0.127 & 0.964 & -- & -- & -- & -- \\
MDETR~\cite{Mdetr} & 0.049 & 0.024 & -- & -- & -- & -- \\
\midrule
\multicolumn{7}{c}{\textbf{MLLM}} \\
\cmidrule(lr){1-7}
Qwen3.6-27B~\cite{qwen36_27b} & \textbf{0.205} & 0.030 & 0.175 & 0.872 & \textbf{0.585} & 0.828 \\
GLM-4.6V-Flash~\cite{GLM-4.6V} & 0.131 & 0.076 & 0.581 & 0.003 & 0.484 & 0.060 \\
MiMo-VL-7B-RL~\cite{MiMo-VL} & 0.047 & 0.000 & 0.105 & 0.000 & 0.095 & 0.000 \\
InternVL3.5-38B~\cite{InternVL35} & 0.148 & 0.006 & 0.441 & 0.147 & 0.384 & 0.233 \\
GLM-4.6V-FP8~\cite{GLM-4.6V} & 0.186 & 0.122 & \textbf{0.611} & 0.000 & 0.505 & 0.087 \\
Gemma-4-31B-it~\cite{Gemma4} & 0.204 & 0.840 & 0.553 & 0.375 & 0.462 & 0.837 \\
\midrule
\multicolumn{7}{c}{\textbf{Grounding-oriented MLLM}} \\
\cmidrule(lr){1-7}
LLaVA-Grounding~\cite{LLaVA-Grounding} & 0.123 & 0.164 & 0.158 & 0.427 & 0.190 & 0.204 \\
GLaMM~\cite{GLaMM} & 0.098 & 0.203 & 0.094 & 0.024 & 0.210 & 0.078 \\
Ferret-v2~\cite{Ferret-v2} & 0.129 & 0.536 & 0.077 & 1.000 & 0.060 & 1.000 \\
KOSMOS-2~\cite{Kosmos-2} & 0.097 & 0.000 & -- & -- & -- & -- \\
Griffon~\cite{Griffon} & 0.011 & 0.011 & 0.061 & 0.741 & 0.040 & 0.628 \\
Rex-Omni~\cite{Rex-Omni} & 0.121 & 0.185 & 0.267 & 0.028 & 0.278 & 0.042 \\
Rex-Seek~\cite{Rex-Seek} & 0.142 & 0.006 & 0.451 & 0.028 & 0.409 & 0.000 \\
LocateAnything-3B~\cite{LocateAnything} & 0.165 & 0.437 & 0.018 & 0.018 & 0.258 & 0.366 \\
\bottomrule
\end{tabular*}
\par\vskip 0.5em
\begin{minipage}{\textwidth}
\footnotesize\raggedright
\emph{Note:} ``--'' indicates that the model does not support the corresponding input modality in that query setting.
\end{minipage}
\end{table*}

We summarize the main observations as follows:

\begingroup
\setlength{\topsep}{2pt}
\setlength{\partopsep}{0pt}
\begin{enumerate}
\item \textbf{2+1 architectures are single-region grounding models.}
TransVG~\cite{TransVG}, CLIP-VG~\cite{clip-vg}, and SimVG~\cite{ming2024simvg} are all formulated for grounding one referred region per query under the standard REC protocol. As a result, they do not natively model variable-cardinality outputs or no-target cases in our benchmark.

\item \textbf{Detection-style models do not natively support image-only queries in our benchmark.} Although detection-style models can in principle output a variable number of boxes, current implementations are primarily text-conditioned and trained outside the UAV perspective. As a result, they remain weak under aerial imagery and do not natively support image-only queries.

\item \textbf{Grounding-oriented MLLMs are weak at precise localization in our setting.} Most of the evaluated grounding-oriented MLLMs are mostly trained under the classic REC regime (text-only, single-target) and lack supervision for multi-instance outputs, no-target discrimination, and multimodal queries, leading to poor localization on our benchmark.

\item \textbf{General-purpose MLLMs transfer better across query modalities, but at a higher cost.} Models such as GLM-4.6V~\cite{GLM-4.6V} and Qwen3.6-27B~\cite{qwen36_27b} adapt more effectively to different query forms and outperform many grounding-oriented MLLMs, suggesting the benefit of large-scale and diverse multimodal pretraining. However, this advantage comes with substantially higher inference cost, which limits deployment efficiency and reproducibility.
\end{enumerate}
\endgroup

Overall, the results in Table~\ref{tab:main-existing-methods} show that Universal Referring is challenging along both input modality and output structure, and that current detection-based and MLLM-based grounding approaches are not yet fully adequate for this setting, indicating the need for domain-specific fine-tuning and architectural adaptation.

\subsection{Baseline Fine-Tuning Results}

To better contextualize the task-specific behavior of UAV-URNet, we include a lightweight general-purpose MLLM as an auxiliary reference by fully fine-tuning Qwen3.5-0.8b~\cite{qwen3.5} on UniRef-UAV, with the detailed training configuration provided in the \emph{Implementation and Evaluation Details} section of the supplementary material. Qwen3.5-0.8b is trained on the merged training split containing all three query forms and evaluated on text-only, image-only, text+image, and merged test splits. In contrast, UAV-URNet is reported under both the merged-training protocol and the modality-matched training protocol. The results are shown in Table~\ref{tab:qwen08b-uavurnet-finetuning}.

\begin{table*}[!t]
\centering
\footnotesize
\setlength{\tabcolsep}{3pt}
\renewcommand{\arraystretch}{1.08}
\caption{Fine-tuning comparison between Qwen3.5-0.8b and UAV-URNet under different test settings.}
\label{tab:qwen08b-uavurnet-finetuning}
\begin{tabular*}{\textwidth}{@{\extracolsep{\fill}}lcccccccc@{}}
\toprule
Model & \multicolumn{2}{c}{Text-only} & \multicolumn{2}{c}{Image-only} & \multicolumn{2}{c}{Text+Image} & \multicolumn{2}{c}{Merged} \\
\cmidrule(lr){2-3} \cmidrule(lr){4-5} \cmidrule(lr){6-7} \cmidrule(lr){8-9}
 & \pmetricsym & \nmetricsym & \pmetricsym & \nmetricsym & \pmetricsym & \nmetricsym & \pmetricsym & \nmetricsym \\
\midrule
UAV-URNet (matched train) & 0.518 & 0.975 & 0.582 & 0.299 & 0.546 & 0.935 & -- & -- \\
UAV-URNet (merged train) & 0.528 & 0.964 & 0.691 & 0.254 & 0.590 & 0.871 & 0.572 & 0.871 \\
Qwen3.5-0.8b~\cite{qwen3.5} (merged train) & 0.320 & 0.492 & 0.772 & 0.324 & 0.675 & 0.340 & 0.485 & 0.456 \\
\bottomrule
\end{tabular*}
\end{table*}

Compared with the fine-tuned Qwen3.5-0.8b, UAV-URNet achieves higher \pmetric{} on the text-only and merged test splits, and consistently obtains much stronger \nmetric{} across all settings. This suggests that a task-specific expert model is more reliable for output-cardinality control and no-target discrimination. In contrast, Qwen3.5-0.8b achieves higher \pmetric{} under image-only and text+image settings, showing the potential of general-purpose MLLMs to exploit visual query cues for target-present cases. However, its substantially lower \nmetric{} indicates a tendency to over-predict boxes and difficulty in recognizing absent targets. These results reveal a clear trade-off: UAV-URNet better fits the structured no-target and variable-cardinality requirements of Universal Referring, while Qwen3.5-0.8b shows stronger positive-sample visual matching in some visual-query settings.

Modality-matched training and merged training exhibit different generalization behaviors. Matched training mainly benefits the corresponding test modality and yields higher \nmetric{} in the text-only and text+image settings, indicating that modality-specific supervision can improve no-target discrimination under matched evaluation. However, it provides limited cross-modal sharing. By contrast, merged training improves \pmetric{} across text-only, image-only, and text+image tests, and directly supports the merged test split. This indicates that different query forms provide complementary supervision and help the model learn a more unified query--target alignment space. For Universal Referring, where one model is expected to handle heterogeneous queries, merged training provides a more practical balance among localization accuracy, modality coverage, and model reuse.

\subsection{Cross-Domain and In-Domain Visual Query Analysis.}

To further analyze the domain generalization ability of visual queries, we evaluate the merged-trained UAV-URNet on image-only and text+image samples that contain visual query inputs. The results are reported in Table~\ref{tab:uav-urnet-domain-eval}. In-domain visual queries consistently yield better localization than cross-domain visual queries, indicating that visual prototypes are easier to match when the query and target scene share a similar domain. Adding text to the visual query improves no-target discrimination, suggesting that textual semantics provide a useful existence constraint. Overall, cross-domain visual queries remain a meaningful generalization challenge in UniRef-UAV: multimodal queries help suppress false positives, but they do not fully remove the localization degradation caused by cross-domain visual prototypes.

\begin{table}[!t]
\centering
\footnotesize
\setlength{\tabcolsep}{4pt}
\renewcommand{\arraystretch}{1.08}
\caption{Cross-domain and in-domain evaluation of UAV-URNet under visual-query settings.}
\label{tab:uav-urnet-domain-eval}
\begin{tabular*}{\columnwidth}{@{\extracolsep{\fill}}llrr@{}}
\toprule
Test setting & Visual query domain & \pmetricsym & \nmetricsym \\
\midrule
\multirow{2}{*}{Image-only} & cross-domain & 0.582 & 0.259 \\
 & in-domain & 0.753 & 0.251 \\
\multirow{2}{*}{Text+Image} & cross-domain & 0.472 & 0.883 \\
 & in-domain & 0.653 & 0.863 \\
\bottomrule
\end{tabular*}
\end{table}

\subsection{Common query embedding space.}

To examine whether merged training induces a shared representation rather than simply memorizing separate query modalities, we conduct a modality-centered similarity analysis on the decoder-head query features; the full protocol and results are provided in the \emph{Common Query Embedding Space} section of the supplementary material. The analysis shows that paired queries across different modalities are consistently more similar than shifted nonpaired queries after modality-wise centering and normalization, with the strongest alignment observed between image-only and text+image queries. These results indicate that UAV-URNet maps different query forms into a comparable query space before detection, supporting a unified query--target alignment interface.

\subsection{Ablation Study}

We finally ablate key design choices in our query formulation.

\textbf{Encoder sharing and freezing}.
Table~\ref{tab:ablation-freeze-share} compares different sharing/freezing strategies in the text+image setting.
The three strategies are numerically close, but they reveal different trade-offs. The freeze+share setting achieves the best \pmetric{} value (0.546), indicating the strongest localization behavior under the text+image setting. Unfreezing the shared encoder slightly improves \nmetric{} (0.945) but reduces \pmetric{} (0.531). The no-share variant does not outperform the shared alternative on both metrics, so we retain \emph{freeze + share} as a compact and stable default.

\begin{table}[htbp]
\centering
\small
\setlength{\tabcolsep}{6pt}
\renewcommand{\arraystretch}{1.1}
\caption{Ablation on sharing/freezing strategy (text+image setting).}
\label{tab:ablation-freeze-share}
\begin{tabular}{cccc}
\toprule
Freeze Weights & Share Weights & \pmetricsym & \nmetricsym \\
\midrule
$\checkmark$ & $\checkmark$ & 0.546 & 0.935 \\
$\times$ & $\times$ & 0.534 & 0.940 \\
$\times$ & $\checkmark$ & 0.531 & 0.945 \\
\bottomrule
\end{tabular}
\vspace{-0.6em}
\end{table}

\begin{table}[!t]
\centering
\small
\setlength{\tabcolsep}{4pt}
\renewcommand{\arraystretch}{1.1}
\caption{Effect of adding text supervision to image-only data.}
\label{tab:ablation-image-only-plus-text}
\begin{tabular}{lcc}
\toprule
Setting & \pmetric & \nmetric \\
\midrule
Image-only & 0.582 & 0.299 \\
Image-only + text query & 0.618 & 0.840 \\
\bottomrule
\end{tabular}
\end{table}

\begin{table}[!t]
\centering
\small
\setlength{\tabcolsep}{4pt}
\renewcommand{\arraystretch}{1.1}
\caption{Effect of removing image prompt from text+image queries.}
\label{tab:ablation-text-plus-image-without-image}
\begin{tabular}{lcc}
\toprule
Setting & \pmetric & \nmetric \\
\midrule
Text+image & 0.546 & 0.935 \\
Text+image w/o image & 0.464 & 0.510 \\
\bottomrule
\end{tabular}
\end{table}

\textbf{Effect of multimodal cues}.
We further isolate the effect of multimodal cues by: (i) augmenting image-only data with text queries, and (ii) removing the image prompt from text+image queries.

As shown in Table~\ref{tab:ablation-image-only-plus-text}, adding text supervision to image-only data raises \pmetric{} from 0.582 to 0.618 and boosts \nmetric{} from 0.299 to 0.840, showing that textual cues help the model handle ambiguous matches and no-target cases more reliably. Conversely, Table~\ref{tab:ablation-text-plus-image-without-image} shows that removing the image prompt from text+image queries degrades both \pmetric{} (0.546 $\rightarrow$ 0.464) and \nmetric{} (0.935 $\rightarrow$ 0.510), confirming that the visual prompt provides complementary grounding information that text alone cannot replace. Taken together, these ablations show that query-side modality design is a substantive factor in Universal Referring: text and image cues are complementary, and the proposed structured query formulation yields more robust behavior across target and no-target cases.

\section{Conclusion}

This paper studies flexible referring understanding for UAV imagery and introduces \emph{Universal Referring}, a task that jointly generalizes the query modality and the output cardinality of conventional REC. To support this setting, we construct UniRef-UAV, a multimodal benchmark covering text-only, image-only, and text+image queries with modality-dependent target cardinality. We further provide UAV-URNet as a detection-style baseline that maps heterogeneous queries into a shared representation space and predicts variable-size target sets through set prediction.

Experiments show that existing grounding models and general-purpose MLLMs are not yet sufficient for this setting, especially under multimodal queries, variable target cardinality, and no-target cases in UAV scenes. UAV-URNet offers a stronger task-specific baseline, with more reliable no-target discrimination and a reproducible implementation. Nevertheless, cross-domain visual queries, precise localization of small aerial objects, and hard no-target cases remain open challenges.

Future work will extend UniRef-UAV along three directions. First, automated data construction pipelines with MLLM-assisted query generation, filtering, and verification may reduce annotation cost and improve query diversity. Second, UAV-specialized MLLMs trained on larger aerial corpora may better combine semantic reasoning with fine-grained localization. Third, lightweight and accurate deployment-oriented models are needed to bring Universal Referring closer to real UAV platforms with limited onboard computation.

\clearpage
\setcounter{page}{1}
\setcounter{section}{0}
\setcounter{subsection}{0}
\setcounter{figure}{0}
\setcounter{table}{0}
\setcounter{equation}{0}
\setcounter{footnote}{0}
\markboth{Supplementary Material}{Supplementary Material}
\twocolumn[{\centering{\Large\bfseries Supplementary Material\par}\vspace{1.5em}}]
\suppressfloats[t]

\section{Data Source Details}

UniRef-UAV is built from images collected from 22 publicly available UAV-related datasets listed in Table~\ref{tab:supp-uav-data-sources}. These sources cover a wide range of aerial scenarios, including urban traffic monitoring, low-altitude flight, and open-field environments, providing diverse visual content for the proposed universal referring setting.

For video-based datasets, we first sample frames at fixed temporal intervals and then merge them with images from static-image datasets. From the merged pool, we uniformly subsample 65,000 images. To reduce redundancy, we further perform visual-similarity-based deduplication to remove near-duplicate frames. During manual annotation, images that are severely blurred or difficult to recognize are discarded. After filtering, UniRef-UAV contains 48,470 unique images.

\begin{table*}[!t]
\centering
\scriptsize
\setlength{\tabcolsep}{2pt}
\renewcommand{\arraystretch}{1.08}
\caption{Summary of UAV-related data sources used in UniRef-UAV.}
\label{tab:supp-uav-data-sources}
\begin{tabular}{@{}L{2.70cm}L{1.30cm}L{3.35cm}L{3.85cm}L{3.40cm}R{1.55cm}@{}}
\toprule
Dataset & Data & Annotation & Target Class & Viewpoint & Images \\
\midrule
AUAIR~\cite{supp:AUAIR} & Image & 2D bbox, tracking/trajectory, scene metadata & Traffic participants: car, bus, truck, pedestrian, bicycle, etc. & UAV top/oblique views over ground scenes & 32,823 \\
DTB70~\cite{supp:DTB70} & Video & Single-object bbox tracking & Generic single objects, including pedestrians, vehicles, boats, and animals & UAV following ground targets & 15,777 \\
DarkTrack2021~\cite{supp:DarkTrack2021} & Video & Single-object bbox tracking & Generic single objects in nighttime scenes & UAV following ground targets & 6,627 \\
ERA~\cite{supp:ERA} & Video & Event-level labels / clip labels & Event and action categories & UAV top/oblique views over ground scenes & 2,864 \\
ITCVD~\cite{supp:ITCVD} & Image & 2D bbox / detection & Vehicles & UAV top views over ground scenes & 173 \\
MDOT~\cite{supp:MDOT} & Video & Multi-object bbox tracking & Vehicles, pedestrians, and other multiple targets & Cooperative multi-UAV top/oblique views over ground scenes & 259,793 \\
Manipal-UAV-Person~\cite{supp:Manipal-UAV-Person} & Image & 2D bbox detection & Person & UAV top/oblique views over ground scenes & 13,462 \\
UAV123~\cite{supp:UAV123} & Video & Single-object bbox tracking & Generic single objects & UAV following ground targets & 113,476 \\
UAVDT~\cite{supp:UAVDT} & Video & 2D bbox detection + MOT + attributes & Vehicles as the main class, with some pedestrian and crowd scenes & UAV top/oblique views over ground scenes & 2,488 \\
UAVDark135~\cite{supp:UAVDark135} & Video & Single-object bbox tracking & Generic single objects in nighttime scenes & UAV following ground targets & 125,466 \\
UAVID~\cite{supp:uavid} & Image & Semantic segmentation masks & Building, road, vegetation, car, human, etc. & UAV top views over ground scenes & 690 \\
UAVVD~\cite{supp:UAVVD} & Video & Rotated boxes / detection + tracking & Vehicles & UAV top views over ground scenes & 1,685 \\
VTUAV~\cite{supp:VTUAV} & Video & Single-object bbox tracking (RGB/thermal-infrared) & Generic single objects & UAV following ground targets & 1,667,629 \\
VisDrone2019~\cite{supp:VisDrone2019} & Image/\allowbreak Video & 2D bbox detection, MOT, SOT & Pedestrian, people, bicycle, car, van, bus, truck, etc. & UAV top/oblique views over ground scenes & 315,519 \\
WebUAV3M~\cite{supp:webuav3m} & Image/\allowbreak Video & Tracking / retrieval-style sequence data & Generic single objects and open-category targets & UAV following ground targets & 753,595 \\
aeroscapes~\cite{supp:aeroscapes} & Image & 2D bbox / segmentation, mainly for aerial semantics & Buildings, roads, vehicles, natural objects, etc. & UAV/low-altitude aerial top views over ground scenes & 6,538 \\
DAC-SDC22~\cite{supp:DAC-SDC22} & Image & 2D bbox detection & Small vehicles / road targets & UAV top views over ground scenes & 93,520 \\
ntut4k~\cite{supp:ntut4k} & Image & 2D bbox detection & Human & UAV top views over ground scenes & 4,095 \\
tacuav~\cite{supp:tacuav} & Image & 2D bbox / traffic object detection & Vehicles and road traffic targets & UAV top views over ground scenes & 16,363 \\
UDD~\cite{supp:udd} & Image & 2D keypoints / sparse annotation / aerial reconstruction related labels & Urban scene objects and building structures & UAV top views over ground scenes & 301 \\
VDD~\cite{supp:vdd} & Image & Semantic segmentation masks & Multiple road and urban scene classes & UAV top views over ground scenes & 400 \\
vsai~\cite{supp:vsai} & Image & 2D bbox detection & Vehicles & UAV top views over ground scenes, multiple viewpoints & 9,075 \\
\bottomrule
\end{tabular}
\end{table*}

\section{Annotation and Split Details}

\subsection{Query Annotation}

We first construct text-only queries. For each image, annotators select either a specific object instance or a group of objects and annotate them with tight bounding boxes. They then write a natural-language description that uniquely specifies the selected target set. Each annotation is further checked to ensure that the textual query is consistent with the annotated boxes in both semantics and target cardinality.

Based on the text-only annotations, we further construct text+image queries by adding visual query examples. We consider two types of visual-query domains: in-domain and cross-domain. For in-domain queries, the visual query and the target image come from the same source environment and correspond to the same object instance. 
Specifically, annotators are provided with a nearby reference frame, sampled within a 2--3 second temporal window around the target image, and are asked to crop the target object from that frame as the image query. The source frame used for the visual query is excluded from the UniRef-UAV target image pool, which preserves instance-level consistency while avoiding image-level duplication between visual queries and dataset images. 
If no suitable reference crop can be found, the sample is retained only as a text-only query.

For cross-domain queries, the visual query and the target image come from different scenes. Annotators search for an image from the web that matches the semantic meaning of the text query, and use it as the visual query. This design allows UniRef-UAV to evaluate both instance-consistent visual grounding and cross-domain visual prototype grounding.
Cross-domain visual queries are intended to serve as semantic visual prototypes rather than instance-consistent references. During annotation, a web-collected query image is accepted only when it matches the core semantic category and key visual attributes specified by the text query, without requiring the same instance, scene, or data domain. For text+image queries, the text defines the semantic boundary and the image provides complementary appearance cues; for image-only queries, we retain only unambiguous zero- or single-target cases to reduce prototype-induced semantic ambiguity.
For web-collected visual queries, we only retain images from copyright-free sources or sources that permit academic redistribution.

\subsection{Output Cardinality Design}

To cover variable target cardinalities, UniRef-UAV explicitly includes no-target queries, where the queried object is absent from the target image despite being plausible in UAV scenarios. This setting reflects practical UAV operations in which a requested target may not appear in the current field of view.

Concretely, we randomly select 15\% of the images for no-target annotation. For each selected image, we first retrieve three candidate queries from the existing annotation pool, prioritizing queries from the same source dataset. A candidate is accepted only when its referred target is semantically absent from the selected target image. If none of the retrieved candidates satisfies this requirement, annotators manually construct a text+image query describing an object that could reasonably appear in the scene but is not present, with the corresponding image query collected from the web. Finally, each no-target query is assigned an in-domain or cross-domain label according to whether the image query and the target image originate from the same dataset.

\subsection{Quality Control}

To ensure annotation quality, we organize the labeling process in multiple batches and apply a mutual verification protocol. In each batch, annotators are divided into two groups, denoted as Group A and Group B, and the data are split accordingly. Each group completes the annotations for its assigned subset and independently reviews 10\% of the annotations produced by the other group. If the error rate identified in the reviewed subset exceeds 1\%, the entire batch annotated by the corresponding group is returned for correction and subsequently rechecked.

\subsection{Image-only Query Design}

Image-only queries are derived from the text+image annotations by removing the textual condition. Accordingly, they follow the same training and testing splits as the text+image query setting. As discussed in the main manuscript, visual queries alone may introduce ambiguity when multiple targets satisfy the same appearance cue. Therefore, for the image-only setting, we exclude samples with multi-target annotations and retain only cases with unambiguous target cardinality.

\subsection{Data Splitting Strategy}

A key requirement in constructing the dataset splits is to preserve the query distribution while preventing image-level leakage across training and testing. We adopt a train/validation/test ratio of 7:1:2 and perform the split under the following constraints:
\begin{enumerate}
    \item Each target image, together with all its associated queries, is treated as the minimum sampling unit.
    \item The same target image is assigned to only one split.
    \item The split ratio is maintained within each source dataset to preserve source-level diversity.
    \item Within each source dataset, the distributions of text-only and text+image queries are kept as balanced as possible.
    \item Within each source dataset, the target-cardinality buckets, i.e., zero, one, and many targets, are also balanced as much as possible.
\end{enumerate}

Based on these principles, we use a source-wise stratified splitting strategy. For each source dataset, target images are first divided according to the predefined ratio. Since different target images may contain different numbers and types of queries, the initial split may still lead to imbalanced query modalities or target-cardinality distributions. We therefore iteratively exchange target-image units between splits to reduce this imbalance, with priority given to the test set, followed by the validation set and then the training set. The procedure stops when the above constraints are satisfied or no further improvement can be obtained.

\section{Implementation and Evaluation Details}
We implement UAV-URNet on top of Grounding-DINO~\cite{supp:GroundingDINO} using the 2+2 detection-style architecture described in the main manuscript. The implementation is based on MMDetection~\cite{supp:mmdetection}. We initialize from pretrained weights of MMGroundingDINO-tiny~\cite{supp:zhao2024open} and fine-tune for 5 epochs with a batch size of 4 per GPU on $2\times$ RTX 4090 GPUs. UAV-URNet is optimized with AdamW using a learning rate of $1\times10^{-4}$ and a weight decay of $1\times10^{-4}$, with gradient clipping at a maximum norm of 0.1. We adopt a MultiStepLR scheduler with milestone at epoch 3, decay factor 0.1, and end epoch 5. On the query side, we adopt the unified representation strategy from the main manuscript, supporting text-only, image-only, and text+image queries with minimal changes to the original detector. We follow the GREC protocol and report \pmetric{} and \nmetric. The score threshold of 0.6 is selected according to validation-set performance and is kept fixed for UAV-URNet, query modalities, and test settings.

For the training and evaluation of multimodal large language models (MLLMs), models with up to 2 billion parameters are deployed and evaluated on an RTX 4090 GPU. Models exceeding 2 billion parameters are evaluated on two RTX PRO 6000 GPUs, with vLLM~\cite{supp:vllm} adopted for efficient inference. As an auxiliary reference for the fine-tuning analysis, we also perform full-parameter fine-tuning of Qwen3.5-0.8b~\cite{supp:qwen3.5} on UniRef-UAV using two RTX PRO 6000 GPUs and compare it with UAV-URNet in the main manuscript. Qwen3.5-0.8b is trained for 3 epochs: the first 2 epochs use the augmented training split, and the final epoch switches to the clean training split. We use a per-device training batch size of 2 and gradient accumulation of 4 on two GPUs, resulting in an effective global batch size of 16.

\subsection{MLLM Prompting and Post-processing Protocol}
For all MLLM baselines, we use model-specific grounding templates to adapt the prompt to each model's native grounding interface. The prompts follow the modality-dependent cardinality of UniRef-UAV: text-only and text+image queries may return a set of matching boxes, whereas image-only queries return at most one box. The Qwen3.6-27B zero-shot baseline uses the same Qwen-style \texttt{bbox\_2d} template as the Qwen3.5-0.8B fine-tuning pipeline, while MiMo-VL follows a Qwen2.5-VL-style \texttt{bbox\_2d} interface with \texttt{/no\_think}. The output format is parsed according to each model's original response convention; for example, Gemma-style \texttt{box\_2d} responses are interpreted in the documented \texttt{[y1, x1, y2, x2]} order before conversion to the unified \texttt{xyxy} format. If the model output cannot be parsed into a valid bounding-box set, we treat it as an empty prediction. This prompting and post-processing protocol is kept fixed across all methods and query settings to ensure fair and reproducible evaluation. The complete model-specific prompt templates and parsing rules will be released together with the evaluation code. Table~\ref{tab:supp-mllm-prompt-examples} gives representative prompt examples for Qwen-style and Gemma-style grounding interfaces.

\refstepcounter{table}\label{tab:supp-mllm-prompt-examples}
\begin{center}
\scriptsize
\setlength{\tabcolsep}{2pt}
\renewcommand{\arraystretch}{1.08}
\textbf{TABLE \thetable}\\
\textbf{REPRESENTATIVE MLLM PROMPT EXAMPLES.}
\vspace{0.45em}
\begin{tabular}{@{}L{1.00cm}L{1.15cm}L{5.60cm}@{}}
\toprule
Model style & Query form & Prompt example \\
\midrule
Qwen-style & text-only & Locate ``\{query\}'' in the image. Output JSON only as a list like [\{\texttt{``bbox\_2d''}: [x1, y1, x2, y2], \texttt{``label''}: ``\{query\}''\}] or \texttt{[]}. Use normalized relative coordinates scaled to 1000. \\
Qwen-style & image-only & You are given two images in order. Image-1 is the full scene. Image-2 is the reference object. Locate in Image-1 the most similar object of the same category as Image-2. Output JSON only as a list like [\{\texttt{``bbox\_2d''}: [x1, y1, x2, y2], \texttt{``label''}: ``target''\}] or \texttt{[]}. Use normalized relative coordinates scaled to 1000. Return at most one box. If no similar object of the same category exists, return \texttt{[]}. \\
Qwen-style & text+image & You are given two images in order. Image-1 is the full scene. Image-2 is the reference object. Locate in Image-1 the object matching Image-2 and the description ``\{query\}''. Output JSON only as a list like [\{\texttt{``bbox\_2d''}: [x1, y1, x2, y2], \texttt{``label''}: ``\{query\}''\}] or \texttt{[]}. Use normalized relative coordinates scaled to 1000. \\
Gemma-style & text-only & What are the bounding boxes for all objects matching ``\{query\}'' in the image? Respond with JSON only as a list like [\{\texttt{``box\_2d''}: [y1, x1, y2, x2], \texttt{``label''}: ``\{query\}''\}] or \texttt{[]}. Coordinates refer to an image size of 1000$\times$1000, relative to the input dimensions. \\
Gemma-style & image-only & Image-1 is the full scene. Image-2 is the visual reference crop. What's the bounding box in Image-1 for the object most similar to Image-2 and of the same category? Respond with JSON only as a list like [\{\texttt{``box\_2d''}: [y1, x1, y2, x2], \texttt{``label''}: ``target''\}] or \texttt{[]}. Coordinates refer to an image size of 1000$\times$1000, relative to the input dimensions. Return at most one box. If no similar same-category object exists, return \texttt{[]}. \\
Gemma-style & text+image & Image-1 is the full scene. Image-2 is the visual reference crop. What are the bounding boxes in Image-1 for all objects that match Image-2 and ``\{query\}''? Respond with JSON only as a list like [\{\texttt{``box\_2d''}: [y1, x1, y2, x2], \texttt{``label''}: ``\{query\}''\}] or \texttt{[]}. Coordinates refer to an image size of 1000$\times$1000, relative to the input dimensions. \\
\bottomrule
\end{tabular}
\end{center}

\section{Common Query Embedding Space}
The main manuscript reports the key finding that UAV-URNet maps text-only, image-only, and text+image queries into a comparable representation space before detection. Here we provide the full analysis protocol and quantitative results.

We use the query features fed into the decoder head as the analysis target. All three query forms are extracted under the same target scene image to avoid additional confounding from scene variation. Let $q_i^{m}\in\mathbb{R}^{d}$ denote the query representation of the $i$-th test sample under modality $m\in\{\mathrm{T},\mathrm{V},\mathrm{TV}\}$, where $\mathrm{T}$, $\mathrm{V}$, and $\mathrm{TV}$ correspond to text-only, image-only, and text+image, respectively. In our implementation, $d=256$. Since raw cosine similarity can be affected by modality-specific mean shifts, we first mean-center and normalize each modality:
\begin{equation}
\tilde{q}_i^{m}=
\frac{q_i^{m}-\mu^{m}}{\left\|q_i^{m}-\mu^{m}\right\|_2},
\qquad
\mu^{m}=\frac{1}{N}\sum_{i=1}^{N}q_i^{m},
\end{equation}
where $N=5000$ is the number of randomly sampled queries from the test split. For any two query modalities $a$ and $b$, we compare the paired similarity from the same sample with the shifted nonpaired similarity from mismatched samples:
\begin{align}
s_{\mathrm{pair}}(a,b)
&=\frac{1}{N}\sum_{i=1}^{N}
\left(\tilde{q}_i^{a}\right)^{\top}\tilde{q}_i^{b}, \\
s_{\mathrm{shift}}(a,b)
&=\frac{1}{N}\sum_{i=1}^{N}
\left(\tilde{q}_i^{a}\right)^{\top}\tilde{q}_{\pi(i)}^{b}.
\end{align}
Here $\pi(i)\neq i$ indexes a shifted nonpaired query. If the three query forms are mapped into a shared semantic space, cross-modal representations from the same sample should be closer than those from mismatched samples:
\begin{equation}
\Delta(a,b)=s_{\mathrm{pair}}(a,b)-s_{\mathrm{shift}}(a,b)>0.
\end{equation}

All three modality pairs satisfy this condition on the test split. To avoid losing information by reporting only the gap, Table~\ref{tab:supp-common-query-space} presents the paired similarity, the shifted nonpaired similarity, and their difference.

\refstepcounter{table}\label{tab:supp-common-query-space}
\begin{center}
\footnotesize
\setlength{\tabcolsep}{2.2pt}
\renewcommand{\arraystretch}{1.08}
\textbf{TABLE \thetable}\\
\textbf{MODALITY-CENTERED QUERY SIMILARITY ON THE TEST SPLIT.}
\vspace{0.45em}
\begin{tabular}{lcccc}
\toprule
Pair & Paired & Shifted & Gap & Paired $>$ Shifted \\
\midrule
T $\leftrightarrow$ V & 0.2271 & -0.0012 & 0.2283 & 73.5\% \\
T $\leftrightarrow$ TV & 0.2572 & -0.0025 & 0.2598 & 66.4\% \\
V $\leftrightarrow$ TV & \textbf{0.6457} & -0.0024 & \textbf{0.6481} & \textbf{94.9\%} \\
\bottomrule
\end{tabular}
\par\vspace{0.5em}
\begin{minipage}{0.96\columnwidth}
\footnotesize \textit{Note:} T, V, and TV denote text-only, image-only, and text+image, respectively.
\end{minipage}
\end{center}

The paired similarities are consistently higher than the corresponding shifted nonpaired similarities, while the shifted values remain close to zero ($-0.0012$, $-0.0025$, and $-0.0024$). This indicates that mismatched queries only exhibit background-level similarity after removing modality-specific mean bias. The strongest evidence comes from the alignment between image-only and text+image queries: $s_{\mathrm{pair}}(\mathrm{V},\mathrm{TV})=0.6457$, compared with $s_{\mathrm{shift}}(\mathrm{V},\mathrm{TV})=-0.0024$. Moreover, $94.9\%$ of paired samples have higher similarity than shifted samples. These results show that the visual representation and the multimodal representation of the same query are close in the decoder/head input space. Therefore, the unified query representation does not simply rely on modality-specific shortcuts. Instead, by mapping text-only, image-only, and text+image queries into the same detection interface, UAV-URNet learns a comparable and shareable common query embedding space, leading to a more unified query--target alignment.

\end{document}